\documentclass[11pt,a4paper]{article}
\usepackage[hyperref]{acl2018}
\usepackage{times}
\usepackage{latexsym}

\usepackage[english]{babel}
\usepackage{graphicx}
\usepackage{amsmath}
\usepackage{amssymb}
\usepackage{tikz}
\usetikzlibrary{positioning,arrows.meta,calc,shapes,fit}
\usepackage[]{subcaption}
\usepackage{booktabs}
\usepackage{nicefrac}
\usepackage{bm}
\usepackage{mathtools}
\usepackage{hyperref}
\usepackage{placeins}
\newcommand{\R}{\mathbb{R}}

\renewcommand{\vec}[1]{{\mathbf #1}}
\newcommand{\x}{\vec x}
\newcommand{\w}{\vec w}
\newcommand{\z}{\vec z}
\newcommand{\h}{\vec h}
\DeclarePairedDelimiterX{\infdivx}[2]{[}{]}{%
  #1\;\delimsize\|\;#2%
}
\newcommand{\infdiv}{D_\text{KL}\infdivx}
\aclfinalcopy % Uncomment this line for the final submission
\title{Explaining Away Syntactic Structure in\\
Semantic Document Representations}

\author{Erik Holmer\thanks{\, Authors contributed equally. Correspondence to {\tt amarfurt@inf.ethz.ch}.} \\
  Department of Computer Science \\
  ETH Z{\"u}rich \\\And
  Andreas Marfurt\footnotemark[1] \\
  Department of Computer Science \\
  ETH Z{\"u}rich}

\date{}

\begin{document}

\maketitle

\begin{abstract}
  Most generative document models act on bag-of-words input in an attempt to focus on the semantic content and thereby partially forego syntactic information. We argue that it is preferable to keep the original word order intact and explicitly account for the syntactic structure instead. We propose an extension to the Neural Variational Document Model \cite{miao2016neural} that does exactly that to separate local (syntactic) context from the global (semantic) representation of the document. Our model builds on the variational autoencoder framework to define a generative document model based on next-word prediction. We name our approach Sequence-Aware Variational Autoencoder since in contrast to its predecessor, it operates on the true input sequence. In a series of experiments we observe stronger topicality of the learned representations as well as increased robustness to syntactic noise in our training data.

\end{abstract}

\section{Introduction}
In natural language processing (NLP), it is becoming increasingly important to be able to classify and cluster different collections of text. Document models are fundamental for many use cases and applications in information retrieval and text mining, for instance, in search ranking and text categorization. They are also often used for collaborative filtering, by making appropriate analogies, e.g. identifying users with documents and items with words.

Most current document models operate on bag-of-words input. On the one hand this is due to computational reasons, but on the other hand there is also the underlying assumption that the removal of the sequence information puts a stronger focus on the topicality of a document. We argue that keeping the input word order intact is in fact preferable and that through explicit modeling of the local syntactic structure, the syntactic and semantic properties of text can be teased apart more effectively.

In this paper, we propose an extension to the Neural Variational Document Model (NVDM) \cite{miao2016neural}, an unsupervised generative document model for text documents of arbitrary length. Our approach employs the next-word prediction model in the variational autoencoder framework \cite{kingma2014auto, rezende2014stochastic}. By explicitly modeling syntax in a local context, our method can separate and "explain away" these effects in the global document representation. Embeddings learned by our model show higher similarity within topics and larger separation between them, and are more robust to syntactic noise in the text datasets of our evaluations. We name our model Sequence-Aware Variational Autoencoder (SAVAE) because in contrast to its base model NVDM, it explicitly incorporates sequence information.

The separation of local and global embeddings in SAVAE is inspired by the Paragraph Vectors model \cite{le2014distributed}, where a paragraph's embedding is concatenated with word vectors from a given context and then used to predict the following word. While providing a simple and powerful way of producing document representations, the paragraph vectors have to be trained at prediction time for any new document. Our model features a proper probabilistic model over documents which lets it infer document representations in a single feedforward pass at prediction time.

In the following, we analyze the state of the art in generative document modeling and highlight the connections between the models. We then introduce our own model which we subsequently evaluate on document retrieval, clustering and sentiment classification tasks. We also inspect the produced word embeddings, and discover that SAVAE nicely separates semantic and syntactic aspects of text.

\section{Related Work}
Generative document models aim to learn a $\theta$-parameterized distribution over words $p_\theta(\w|l)$ for a given document $\w$ of length $l$,
\begin{align*}
    \w = w^1, \ldots, w^l, \quad w^t \in \mathcal{V} = \{w_1, \ldots, w_m\},
\end{align*}
where $\mathcal{V}$ is a predefined vocabulary of size $m$. In order to focus on the topicality of a document, the distribution is often over word multisets $\vec n(\w)$ instead of the exact sequence $\w$, which is also called the bag-of-words model.

The introduction of topic models has drawn a lot of attention to the field. Latent Dirichlet Allocation (LDA) \cite{blei2003latent} is a particularly renowned topic model where documents are viewed as mixtures over latent topics. These topics define a distribution over words. In the specified generative process, for each word in the document a topic is sampled independently. Subsequently, the word itself is sampled according to the distribution of that topic.

The Replicated Softmax \cite{hinton2009replicated} is an adaptation of the general Restricted Boltzmann Machine to the bag-of-words model. It consists of a layer of visible units that represent the input bag-of-words, and a layer of hidden binary units that can be seen as topics and collectively serve as the document representation.

A common definition of models for arbitrary length word sequences exploits the chain rule
\begin{align}
    p_\theta(\w | l) = \prod_{t=1}^{l} p_\theta(w^{t} | \w^{1:t-1}).
\end{align}
All we need to provide is a next-word prediction model, which is typically given in the form of a softmax function
\begin{align}
    p_\theta(w|\w) = \frac{\exp \left[ \x_w^\top \z_\w + b_w \right]}{\sum_{v \in \mathcal{V}} \exp \left[ \x_v^\top \z_\w + b_v \right]}\,,
    \label{eq:softmax}
\end{align}
where $\x_v \in \R^d$ are the word embeddings, $b_v \in \R$ are biases, and $\z_\w$ are word sequence embeddings computed from the context $\w$.
A proponent of the next-word model based on the Neural Autoregressive Distribution Estimator (NADE) \cite{larochelle2011neural} is Document NADE (DocNADE) \cite{larochelle2012neural}. To compute the context embeddings, it uses 
\begin{align}
    \z_\w = \sigma \left(\mathbf c + \sum_{w \in \w}\z_{w}\right),
    \label{docnade:context}
\end{align}
where $\sigma$ is the sigmoid nonlinearity. Therefore, there are two embedding vectors of equal dimension per word in the vocabulary, $\x_w, \z_w \in \R^d$. We can see from Eq.~\eqref{docnade:context} that the context embedding is independent of the word ordering in the context $\w$. Although the next-word model operates on the true word sequence of a document, the authors of DocNADE assume that this ordering is not readily available for many datasets, and they therefore train on randomized word orderings. The paper notes that the performance of the model when trained on two different random orderings is comparable.

In DeepDocNADE \cite{lauly2016document} the DocNADE model is extended to compute the context embeddings $\z_\w$ with a multilayer architecture. It completely removes the autoregressive parts in DocNADE, and instead predicts the remaining words in the document from a bag-of-words context without any word ordering assumptions.

A different approach than the next-word model is the latent Bayesian model that introduces a latent variable $\z \in \R^d$ to define the probabilistic model as
\begin{align}
    p_\theta(\w | l) &= \int p_\theta(\w | \z,l)p(\z)d\z \\
    &= \int\prod_{t=1}^{l}p_\theta(w^t|\z)p(\z)d\z\,,
    \label{nvdm:integral}
\end{align}
where one typically assumes that $p(\z)$ is simple (and not parameterized) and we can use the softmax for $p(w|\z)$ as in Eq.~\eqref{eq:softmax}. The problem with the latent variable model is that we cannot easily perform the integration, even if $p(\z)$ is as simple as an isotropic normal distribution. Hence, we cannot evaluate the model, e.g. to compute a document probability or perplexity.

A solution to this problem has been introduced with the Variational Autoencoder (VAE) \cite{kingma2014auto, rezende2014stochastic}, which approximates the integral by maximizing the variational lower bound. The Neural Variational Document Model \cite{miao2016neural} applies the VAE framework to document modeling. The inference network of the autoencoder estimates the mean and spread of a variational distribution $q(\z|\w)$ over the latent variable $\z$. Specifically, the mean is modeled as
\begin{align}
    \mathbf{E}_q[\z|\w] = \left(\vec c + \sum_{w \in \w} \z_{w} \right)_+,
\end{align}
where $(\cdot)_+$ is the rectified linear activation function. The decoder $p_\theta(w|\z)$ is again the softmax function from Eq.~\eqref{eq:softmax}. From a modeling perspective, NVDM is very similar to DocNADE, with two exceptions: (i) the ReLU nonlinearity was applied to infer context embeddings, and (ii) in addition to the mean a dispersion parameter is estimated for the latent variable $\z$. This effectively adds noise to the training process, as the sampling from $q$ generates slightly different results in each training pass.

Recent work attempted to apply Generative Adversarial Networks (GAN) to model documents \cite{glover2016modeling}. While this seems to be a promising direction of research, the results so far are still behind the state of the art.

Interesting models for shorter sequences of text (sentences) apply RNNs in a VAE framework \cite{bowman2016generating}. The authors report difficulties in making efficient use of the latent representation, most likely due to a too powerful RNN decoder that can independently explain the structure in the data. This is further analyzed in \cite{chen2017variational} where the authors state under which conditions the latent code is used in a VAE with an autoregressive decoder. Namely, when the autoregressive model is restricted to operate on limited local context, the global semantics must be captured in the latent code. This observation is a key element to the design of our model.

Multiple recent studies work on the encoder of VAEs. \cite{kingma2016improved} use inverse autoregressive flows to obtain more flexible latent distributions. Similarly, \cite{serban2017piecewise} employ a piecewise constant prior to allow for multiple modes. The Stein variational autoencoder \cite{pu2017vae} uses Stein's identity to get rid of the prior distribution for the latent variable altogether. These approaches are orthogonal to our method, and it would be interesting to see the results of combining them.

Outside of the realm of generative models, a different method has been proposed to solely learn the document representations. The method was originally named Paragraph Vectors \cite{le2014distributed}, but is also commonly referred to as doc2vec because of its similarities to the word2vec method \cite{mikolov2013efficient} for learning word embeddings. doc2vec exists in two versions, called the distributed memory model (PV-DM) and the distributed bag-of-words model (PV-DBOW). The PV-DM model predicts the next word from a paragraph/document embedding together with word embeddings from the previous words. The document and word embeddings are then updated by backpropagation. This model has strong similarities to our model, as we discuss in Section~\ref{model}. The other model, PV-DBOW, predicts words that are randomly sampled from the current paragraph, solely based on the document embedding. This method completely ignores the local context, but is conceptually simpler and faster to train.

Semantic document representations can benefit other tasks, such as language modeling. As an example, TopicRNN \cite{dieng2017topicrnn} uses the same encoder as NVDM to provide its decoder with additional semantic information. We can imagine that work in our area can be incorporated in recurrent models working on such tasks.

\section{Model}
\label{model}
In \cite{le2014distributed}, the authors propose a novel way of generating low dimensional representations for paragraphs/documents. One of the main ideas in the paper is to maintain a global representation of a document and combine it (through concatenation or averaging) with word vectors from a local context of $k$ previous words to predict the next word. The hope is that this separates global from local features, allowing the model to attend to them independently and thus producing better results.
\begin{figure}[tbp]
\centering
\resizebox{\linewidth}{!}{\tikzset{
  arro/.style={
    ->,
    >=latex
  },
  bloque/.style={
    draw,
  },
  stack/.style={
      rectangle split,
      rectangle split horizontal, 
      rectangle split parts=6, 
      draw, 
      anchor=center,
      inner sep=0.0cm,
  }
}
\begin{tikzpicture}[node distance=1cm and 0.5cm]
\node[]
  (embed)
  {\footnotesize Embeddings $\z_v, \vec h_v$};
\node[stack,below=of embed,label={[name=z_the]above:$\z_{the}$}]
  (embedi)
  {};
\node[stack,below=of embedi,label={above:$\z_{cat}$}]
  (embedii)
  {};
\node[below=of embedii]
    (aux)
    {};
\node[stack,below=of embedii,label={above:$\z_{sat}$}]
  (embediii)
  {};
\node[below=of embediii]
    (aux4)
    {};
\node[stack,below=of embediii,label={above:$\z_{on}$}]
(embediv)
{};

\node[stack,below=of embediv,label={[name=h_the] above:$\vec h_{the}$}]
  (embedv)
  {};
\node[below=of embedv]
    (aux3)
    {};
\node[stack,below=of embedv,label={above:$\vec h_{cat}$}]
  (embedvi)
  {};

\node[left=of embed]
    (input)
    {\footnotesize Input $\w$};
\node[]
    at (input|-aux4)
    (aux2)
    {};
\node[bloque,above=of aux2,label={above:$w^2$}]
    (inputii)
    {cat};
\node[bloque,above=of inputii,label={above:$w^1$}]
    (inputi)
    {the};
\node[bloque,below=of inputii,label={above:$w^3$}]
    (inputiii)
    {sat};
\node[bloque,below=of inputiii,label={above:$w^4$}]
    (inputiv)
    {on};

\node[right=of embed, align=center]
  (trans)
  {\footnotesize Transform};
\node[stack,label={above:$\bm \mu$}]
  at (trans|-embedii)
  (transi)
  {};
\node[stack,label={above:$\bm \sigma$}]
  at (trans|-embediii)
  (transii)
  {};
  
\node[right=of trans, align=center]
  (sample)
  {\footnotesize Sample/Sum};
\node[stack,label={[name=q] above:$\z \sim q(\z|\w)$},yshift=0.5cm]
  at (sample|-aux)
  (samplei)
  {};
\node[stack,label={[name=h_w] above:$\vec h_\w$},yshift=0.5cm]
  at (sample|-aux3)
  (sampleii)
  {};

\node[right=of sample]
  (out)
  {\footnotesize Predict $w^3$};
\node[stack,label={[xshift=0.5cm] above:$p_\theta(w^3|\w^{1:2},\z)$}]
  at (out|-embediv)
  (outi)
  {};
\foreach \Valor in {i,ii,iii, iv}
{
  \draw[arro] (input\Valor) -- (embed\Valor.west);
}
\draw[arro, dotted] (inputi) -- (embedv.west);
\draw[arro, dotted] (inputii) -- (embedvi.west);
\foreach \Valor in {i,ii,iii, iv}
{
  \draw[arro] (embed\Valor.east) -- (transi.west);
  \draw[arro] (embed\Valor.east) -- (transii.west);
}
\draw[arro] (embedv) -- (sampleii);
\draw[arro] (embedvi) -- (sampleii);
\draw[arro] (transi.east) -- (samplei.west);
\draw[arro] (transii.east) -- (samplei.west);
\draw[arro] (samplei.east) -- (outi.west);
\draw[arro] (sampleii.east) -- (outi.west);
\node (local) [draw, fit= (h_the) (embedvi) (sampleii), dashed,yshift=-0.2cm] {};
\node (global) [draw, fit= (z_the) (embediv) (q), dashed] {};
\node [yshift=1.3ex] at (local.south) {\footnotesize Local context};
\node [yshift=1.3ex] at (global.south) {\footnotesize Global context};
\draw[draw=gray] (input.south west) -- (out.south east);
\end{tikzpicture}}
\caption{Illustration of how SAVAE predicts the next word~$w^3$.
We look up global word embeddings $\z_v$ for all words in $\w$ and local embeddings $\vec h_v$ for the $k$ previous words $\w^{1:2}$. 
The global word embeddings are then summed up and transformed through $N$ layers of an MLP, inferring $\bm \mu$ and $\bm \sigma$. Then we sample from $q(\z|\w)$ to obtain a document representation $\z$ which is concatenated with the local embedding $\vec h_\w$ and finally used to predict the next word $w^3$.}
\label{fig:savae_architecture}
\end{figure}
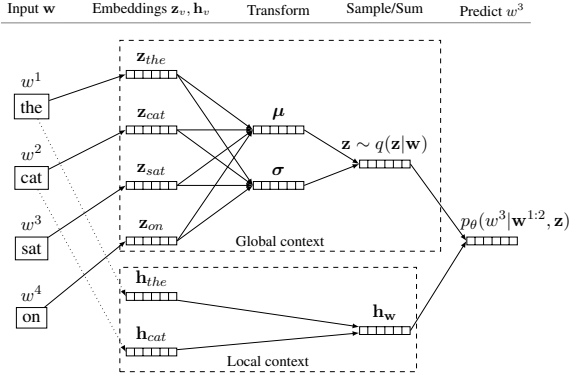
In this section we take this idea of combining local and global context features to build a generative document model.
We start by defining a generative model
\begin{align}
    p_\theta(\w | l) = \int p_\theta(\w | \z,l)p(\z)d\z,
    \label{savae:integral}
\end{align}
with a latent variable $\z \in \R^d$, which we will refer to as the global context vector.
We can interpret the model as a VAE, defining an encoder $q(\z|\w)$ and a decoder $p_\theta(\w | \z, l)$.
We let the decoder reconstruct documents word by word based on the global context vector $\z$ and a small window of the previous $k$ words. Thus Eq.~\eqref{savae:integral} becomes
\begin{align}
    p_\theta(\w | l) = \int\prod_{t=1}^{l}p_\theta(w^t|\w^{(t-k):(t-1)}, \z)p(\z)d\z.
    \label{savae:integral2}
\end{align}
We define the next-word prediction model as the softmax function
\begin{align}
    p_\theta(w|\w, \z) = \frac{\exp\left[\x_w^\top(\z, \h_\w) + b_w\right]}{\sum_{v \in \mathcal V}\exp\left[\x_v^\top(\z, \h_\w) + b_v\right]},
    \label{savae:softmax}
\end{align}
where $b_w$ are biases, $\x_w$ are word embedding vectors taken from a matrix $\vec V_{global} \in \R^{m\times2d}$ and $(\z, \h_\w)$ is the concatenation of the global context vector $\z$ and the local context vector $\h_\w$, which we define as
\begin{align}
    \h_\w = \sigma\left(\vec c + \sum_{w \in \w}\h_{w}\right),\quad \sigma=\text{sigmoid},
\end{align}
where $\vec c \in \R^d$ is a bias vector, and $\h_w \in \R^d$ are word embedding vectors taken from a matrix $\vec V_{local} \in \R^{m\times d}$. Note that within each context $\w^{(t-k):(t-1)}$, the ordering of the words is ignored and their embeddings simply added up.

The encoder distribution $q(\z|\w)$ is modeled as a Gaussian $\mathcal N(\bm \mu,\bm \sigma)$ where the mean and spread are inferred using a standard feedforward MLP of $N$ layers.
Now, instead of directly optimizing Eq.~\eqref{savae:integral2} we can use the variational lower bound on the log-likelihood (ELBO) as our training objective
\begin{align}
\begin{aligned}
    \log\, & p_\theta(\w|l) \geq \\
    &\mathbf E_{q(\z|\w)}\left[\sum_{t=1}^{l} \log p_\theta(w^t|\w^{(t-k):(t-1)}, \z)\right] \\
    &-\infdiv{q(\z|\w)}{p(\z)},
\end{aligned}
\label{savae:elbow}
\end{align}
where $p_\theta(w|\w,\z)$ is the softmax function in Eq.~\eqref{savae:softmax}. In the particular case where $p(\z)$ is a standard Gaussian, the KL divergence term $D_\text{KL} \left[ \cdot \| \cdot \right]$ can be computed analytically, as described in \cite{miao2016neural}. Note that we need to sample $\z$ from $q(\z|\w)$ to approximate the expectation in Eq.~\eqref{savae:elbow}. An illustration of how SAVAE predicts the next word in a document $\w$ is shown in Figure~\ref{fig:savae_architecture}.

After training the model, we can use the ELBO to give a lower bound on the probability for a document $\w$. We can also use the inferred $\bm \mu$ as a representation of a document for further use, e.g. classification, information retrieval or clustering.

\section{Experiments}
We evaluate the semantic document representations and inspect the learned word embeddings. We compare our model against its base version NVDM, which uses only the global context. We further evaluate doc2vec, which obtains document representations in a slightly different manner but also takes the local context into account. Finally, we also analyze DocNADE as a different proponent of the next-word prediction model.

In the following, we first describe the datasets, preprocessing and the choice of parameters that were used in training, before presenting the results of our experiments. Code and data to reproduce the experiments will be made available.

\subsection{Datasets and Preprocessing}
Three standard text corpora were used for the subsequent evaluations. They are 20 Newsgroups, Reuters Corpus Volume 1 version 2 (RCV1-v2) \cite{lewis2004rcv1}, and the IMDB movie review dataset \cite{maas2011learning}. The 20 Newsgroups corpus contains 18,845 posts from Usenet discussion groups. These posts range across 20 diverse discussion topics such as computer hardware, sports and religion, which also serve as their labels. The Reuters RCV1-v2 is an archive of 804,414 newswire stories that were categorized by their editors into possibly multiple topics out of 103 possibilities. These labels are structured in a tree hierarchy, and documents inherit all parent labels of their most specific label. The IMDB movie review dataset contains 100,000 movie reviews, where 25,000 are labeled with a positive sentiment, 25,000 with a negative one, and 50,000 come without a label.

Since our model requires the actual word sequence as input, we cannot use the same preprocessed data that was used in \cite{hinton2009replicated} and later prior work. Therefore, we keep the preprocessing standard and adopt choices from prior work where we can. For all corpora, lowercasing and tokenization of scikit-learn is applied. We do not remove any stopwords. We do not apply any discounting to the word counts. Further corpus-specific preprocessing is stated in the following.
\paragraph{20 Newsgroups.} As in previous work, we restrict the vocabulary to the 2,000 most common words. We use the same train/test split. We then shuffle the respective datasets with a random seed of 2. By default, the 20 Newsgroups posts come with metadata such as headers, footers and quotes that appear in typical email conversations. These content-wise uninformative data can vary strongly between discussion groups, but stay consistent within a group. Thus, they provide strong signals for classifiers and also influence unsupervised representation learning. We consider this an artefact of data collection and not representative of general documents, and therefore remove metadata from the corpus.
\paragraph{Reuters RCV1-v2.} Following prior work, the vocabulary size was set to 10,000. We also randomly split the data into 794,414 training and 10,000 test articles. Note that this is not the same random split.
\paragraph{IMDB movie reviews.} We keep a vocabulary of the 10,000 most common words. The original dataset comes with train and test splits for positive and negative samples, along with unlabeled reviews. Since we learn the document representations completely unsupervised, we pool all the data together to learn the document representations as in \cite{le2014distributed}.

\begin{figure*}[!ht]
\centering
\begin{minipage}{.49\linewidth}
\centering
\includegraphics[width=\linewidth]{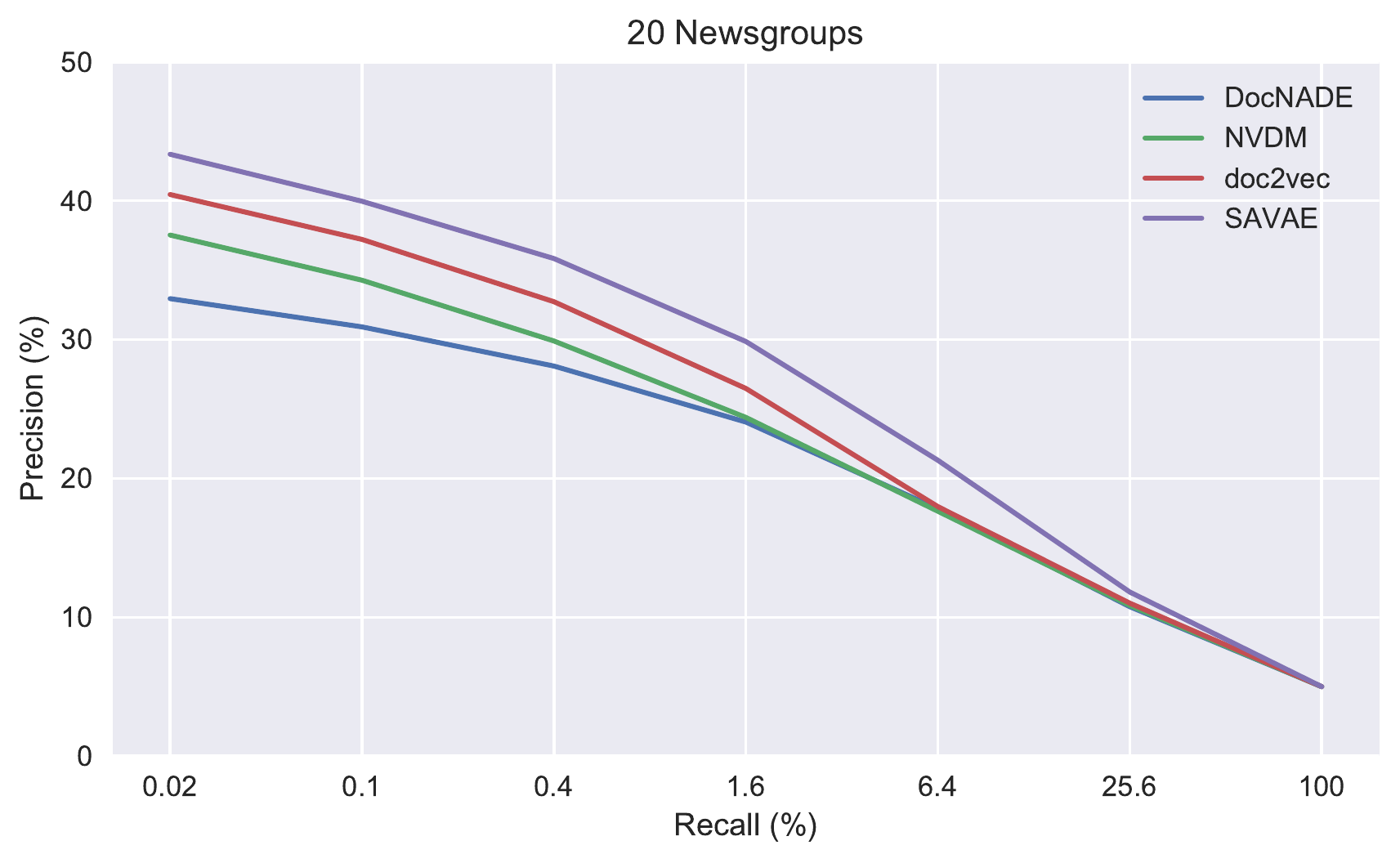}
\end{minipage}
\begin{minipage}{.49\linewidth}
\centering
\includegraphics[width=\linewidth]{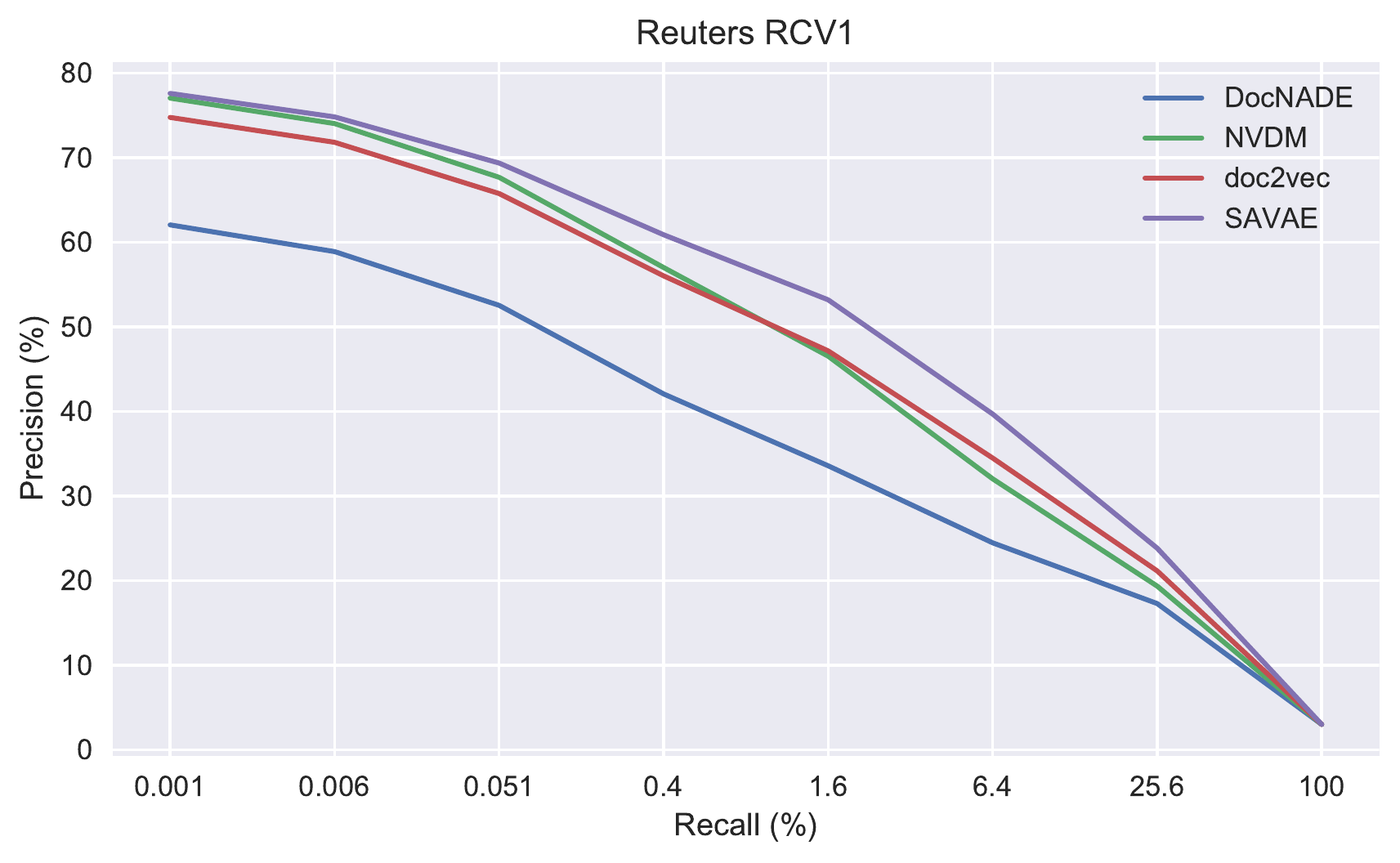}
\end{minipage}
\caption{Document retrieval evaluation on 20 Newsgroups (\textbf{left}) and Reuters RCV1 (\textbf{right}).}
\label{fig:docretrieval}
\end{figure*}

\subsection{Training Details}
We set the dimensionality of our embeddings to 50 for 20 Newsgroups and Reuters, and to 100 for IMDB, following previous work. We trained all generative models for 1,000 epochs on 20 Newsgroups and IMDB, and 100 epochs on Reuters. The exception is doc2vec, which we only trained for 20 epochs on all corpora, as results did not improve when we trained for more epochs. We found that after convergence the results remained stable for all algorithms.

In the course of analyzing the various models, we reimplemented all but doc2vec in the Tensorflow framework. This allowed us to adjust the model parameters, and in the process we found that using the Adam optimizer \cite{kingma2015adam} improved results for all models.
\paragraph{DocNADE.} We used a learning rate of 0.001. Training for DocNADE was rather unstable, so we employed early stopping based on the best attained perplexity during training.
\paragraph{NVDM.} We used a learning rate of 0.0001. The inference network consists of two layers of dimension 500 and ReLU nonlinearities. To approximate the variational lower bound we use 1 sample during training and 20 samples during evaluation, as suggested in \cite{miao2016neural}.
\paragraph{doc2vec.} Since no code was published with the original paper, we used an implementation by the co-author Tomas Mikolov~\footnote{posted in \href{https://groups.google.com/forum/\#!msg/word2vec-toolkit/Q49FIrNOQRo/J6KG8mUj45sJ}{this Google Groups discussion}} to train doc2vec. Although our model is very similar to the distributed memory model (PV-DM) in the original paper, later studies \cite{dai2015document, lau2016empirical} have found the distributed bag-of-words model (PV-DBOW) to perform better. Since we observed the same behavior in our own experiments, we use the distributed bag-of-words version for our evaluations. We jointly train the word and document embeddings. The algorithm uses stochastic gradient descent as an optimizer, with a learning rate that linearly decays from 0.05 to 0. We used a window size of 10, and applied negative sampling with 10 negative samples per positive sample. We downsampled frequent words above a frequency of 0.0001.
\paragraph{SAVAE.} We fixed the learning rate at 0.00001. The inference network is the same as in NVDM. In the decoder, we achieved the best results with $k=5$ for the size of the local context. To approximate the variational lower bound we use 1 sample during training and 20 samples during evaluation, just as in NVDM.

\subsection{Document Retrieval Evaluation}
\label{document-retrieval}
Adopting the document retrieval evaluation from prior work, we unsupervisedly train document representations on the train sets of 20 Newsgroups and Reuters. For each query document in the held-out test set, we first get its document representation while keeping the model parameters fixed. We then rank the training documents by their representations' cosine similarity with the query document's representation. We can then compute precision-recall curves as we go from documents that have been predicted to be similar to those that are supposed to be dissimilar.

How precision is computed differs between the two datasets, since the documents of 20 Newsgroups have only one label (out of 20 categories), whereas the Reuters documents can have multiple labels (out of 103 labels). For 20 Newsgroups we compute the precision at rank $r$ as the fraction of $r$ most similar documents with the same label as the query document. For the Reuters dataset we define the relevance of a document in the training set for a given query document to be the Jaccard similarity between their respective label sets, i.e. the cardinality of the intersection divided by the cardinality of the union of the two sets. For both datasets we get a precision-recall curve per query document that we average for the final results.

The results for 20 Newsgroups can be seen in Figure~\ref{fig:docretrieval} on the left. SAVAE improves substantially on its base model NVDM, and even beats the representations of doc2vec. DocNADE performs the worst throughout the entire task. It is interesting to note that SAVAE's performance does not degrade like the other models' for higher recall rates. On the right-hand side of Figure~\ref{fig:docretrieval}, we show the evaluation on the Reuters RCV1 corpus. For very low recall, NVDM and SAVAE perform almost identical. As we increase the recall, NVDM loses precision and matches the performance of doc2vec. The gap between doc2vec and SAVAE slightly widens as recall increases.

\begin{table*}[!ht]
\centering
\begin{tabular}{lcccc}
\toprule
Model & Davies-Bouldin index $\downarrow$ & Dunn index $\uparrow$ & Silhouette coefficient $\uparrow$ \\
\midrule
DocNADE & 124.0651 $\pm$ 62.0585 & 0.0076 & -0.0117 $\pm$ 0.0566 \\
NVDM & 5.2976 $\pm$ 2.1503 & 0.1905 & 0.0745 $\pm$ 0.0715 \\
doc2vec & 6.5670 $\pm$ 3.0634 & 0.1405 & \textbf{0.0970 $\pm$ 0.0770} \\
SAVAE & \textbf{3.2995 $\pm$ 1.3666} & \textbf{0.2875} & 0.0911 $\pm$ 0.0810 \\ \bottomrule
\end{tabular}
\caption{Clustering metrics on representations trained on 20 Newsgroups. An upwards arrow ($\uparrow$) indicates that a higher score is better, a downwards arrow ($\downarrow$) that lower is better. Where applicable, the mean and standard deviation across clusters is reported.}
\label{tab:cluster-analysis}
\end{table*}

\subsection{Cluster Analysis}
\label{cluster-analysis}
We decide to further analyze the properties of clusters produced from 20 Newsgroups data. While this task was not employed in prior work, we adopt it in order to gain insight into the structure of our document embedding space. We want to know how well the algorithms separate different topics, and how tightly they group related documents.

Our algorithms only output an implicit clustering through the documents' vector representations; in particular no labels are output. We thus resort to internal evaluation metrics which evaluate the intra- and inter-cluster distances. For a robust solution to the clustering problem, we prefer the intra-cluster distances to be low, while the inter-cluster distances should be high. In our evaluation, we use three established metrics that evaluate this principle from slightly different angles.

The Davies-Bouldin index~\cite{davies1979cluster} is defined as
\begin{align}
\frac{1}{n} \sum_{i=1}^n \max_{j \neq i} \left( \frac{\pi_i + \pi_j}{d(c_i, c_j)} \right),
\end{align}
where $n$ is the number of clusters, $\pi_i$ is the mean distance of all elements of cluster $i$ to its centroid $c_i$, and $d$ is our distance measure. The Davies-Bouldin index should be minimized, which can be achieved with low intra-cluster distances ($\pi$) and high inter-cluster  distances ($d(c_i, c_j)$). Note that only the cluster that maximizes this ratio is considered, which therefore is the most similar one.
Second, the Dunn index~\cite{dunn1973fuzzy} is
\begin{align}
\frac{\min_{1 \leq i < j \leq n} d(c_i, c_j)}{\max_{1 \leq k \leq n} \pi_k}.
\end{align}
The Dunn index inverts the ratio of intra- and inter-cluster distances compared to the Davies-Bouldin index, and should consequently be maximized. Furthermore, instead of computing a score per cluster, it outputs a global ratio of the lowest inter-cluster distance to the highest intra-cluster distance. It therefore penalizes the currently worst cluster.
Finally, the silhouette coefficient~\cite{rousseeuw1987silhouettes} computes a point-wise normalized distance to the true cluster centroid and the closest wrong centroid. We further average the per-cluster scores, to balance the contributions of the individual clusters.
\begin{align}
\frac{1}{n} \sum_{i=1}^n \sum_{x \in C_i} \frac{\min_{j \neq i} d(x, c_j) - d(x, c_i)}{\max \{ d(x, c_i), d(x, c_j) \} }
\end{align}
For an individual data point $x$, a score of zero means that the true and the closest wrong cluster's centroid have an equal distance to $x$. A negative score means that there is a centroid closer to $x$ than the true one, and a positive score indicates that the true centroid is also the closest.

We use the cosine distance as our distance measure $d$ and normalize the centroid distances per data point, so that outliers cannot influence the scores overproportionately. For the Davies-Bouldin index and the silhouette coefficient, where we average the cluster scores, we also report the standard deviation across clusters.

The results are shown in Table~\ref{tab:cluster-analysis} and confirm the findings of the previous experiment. While the absolute values have no significance to us, we can qualitatively determine that SAVAE produces the most robust clustering with higher inter- and lower intra-cluster distances than the other methods. The silhouette coefficient is positive for NVDM, doc2vec and SAVAE, which indicates that for the most part, the true centroid is also the closest for a given data point. The variance between clusters is rather large, which hints at the fact that some topics of the 20 Newsgroups dataset are harder to separate.

\begin{table*}[!ht]
\centering
\resizebox{\linewidth}{!}{%
\begin{tabular}{ccc|ccc|ccc|ccc}
\multicolumn{3}{c}{\textbf{NVDM}} & \multicolumn{3}{c}{\textbf{doc2vec}} & \multicolumn{3}{c}{\textbf{SAVAE (global)}} & \multicolumn{3}{c}{\textbf{SAVAE (local)}} \\
\toprule
\textbf{weapons} & \textbf{medical} & \textbf{define} & \textbf{weapons} & \textbf{medical} & \textbf{define} &  \textbf{weapons} & \textbf{medical} & \textbf{define} & \textbf{weapons} & \textbf{medical} & \textbf{define} \\ \midrule
guns             & medicine         & defined         & weapon           & health           & defined         & weapon           & health           & draw            & weapon           & basic            & with            \\ %\midrule
weapon           & disease          & null            & firearms         & disease          & definition      & firearms         & disease          & realize         & practice         & serial           & make            \\ %\midrule
batf             & health           & int             & guns             & patients         & must            & arms             & medicine         & assume          & files            & party            & completely      \\ %\midrule
firearms         & patients         & morality        & defense          & study            & indeed          & guns             & patients         & count           & operation        & graphics         & include         \\ %\midrule
militia          & treatment        & constitution    & citizens         & volume           & false           & crime            & treatment        & notice          & event            & page             & towards         \\ \bottomrule
\end{tabular}%
}
\caption{Learned word embeddings on the 20 Newsgroups corpus for NVDM, doc2vec and SAVAE.}
\label{wemb:comparison}
\end{table*}

\begin{table}
    \begin{center}
        \begin{tabular}{lc}
            \toprule
            Model & Accuracy \\
            \midrule
            DocNADE & 80.79 \% \\
            NVDM & 88.96 \% \\
            doc2vec & 89.89 \% \\
            SAVAE & 89.05 \% \\
            \midrule
            TopicRNN* & 93.72 \% \\
            Virtual Adversarial* & \textbf{94.09 \%} \\
            \bottomrule
        \end{tabular}
    \end{center}
    \caption{Classification accuracy on the sentiment classification task of IMDB movie reviews. Results with an asterisk~(*) are taken from the respective publications.}
    \label{tab:sentiment}
\end{table}

\subsection{Word embeddings}
SAVAE produces semantic (global) and syntactic (local) embeddings. We therefore want to inspect these and compare them to the embeddings produced by our baselines. We select a subset of words chosen in previous studies~\cite{larochelle2012neural, miao2016neural} and show their 5 nearest neighbors by cosine distance in Table~\ref{wemb:comparison}.

We can see that the word embeddings of NVDM, doc2vec and SAVAE (global) are very similar and semantically meaningful for the words \textit{weapons} and \textit{medical}. The biggest difference shows for the word \textit{define}, where both NVDM and doc2vec group the word with a context of program code (\textit{null}, \textit{int}, \textit{false}). This is a syntactic similarity based on co-occurrence instead of a related meaning. SAVAE nicely separates synonyms from co-occurring terms into the global and local embeddings, respectively.

Taking a closer look at the local embeddings of SAVAE, we can see two effects at work. The co-occurrence in many contexts is a straightforward effect that causes local embeddings of two words to move closer together. \textit{define} often co-occurs with adverbs and prepositions, which gets captured accordingly. A second effect we found was the existence of \textit{connector} words that appear in the contexts of two seemingly unrelated words. Examples for such connector words are \textit{organization}, which has a high co-occurrence count with both \textit{medical} and \textit{graphics}, and \textit{research} for \textit{medical} and \textit{basic}. Finally, there are artefacts specific to our data that do not have a linguistic reason. We observe many co-occurrences of \textit{medical} and \textit{page} due to a phrase that looks like the page footer of a scientific journal that did not get filtered out by preprocessing. However, this serves as yet another compelling argument to separate accidental syntactic co-occurrences from the semantic representations of words and documents.

\subsection{Sentiment Classification}
An interesting task adopted in the doc2vec paper \cite{le2014distributed} involves sentiment classification on IMDB movie reviews. In a first step, document representations are trained in an unsupervised manner on all reviews. The representations of 12,500 samples of both positive and negative sentiment are then used to train a linear classifier that predicts the polarity of the 25,000 test samples. In representation learning, a popular view holds that a good representation should be able to "disentangle" the factors of variation present in the data and make them linearly separable. In this task, we test this hypothesis for the dimension of sentiment. The results are listed in Table~\ref{tab:sentiment}. We observe that all the semantic document models perform very similarly, with the exception of DocNADE. To our knowledge, the current state-of-the-art is held by an approach that combines virtual adversarial training with recurrent neural networks~\cite{miyato2017adversarial} and uses much higher dimensional hidden representations. TopicRNN, which uses the same encoder as NVDM, is close behind.

The results are somewhat expected. It is well known that word sequence plays an important role in sentiment classification. This can be demonstrated with the example of negation, where the adverb \textit{not} changes the sentiment of a statement. The cases where syntax decides the sentiment will lead to classification errors for semantic document representations, since they either ignore, or -- in the case of SAVAE and doc2vec -- explicitly explain away the syntactic structure. It is remarkable, however, how close to the state of the art the document representations get considering they are not trained end-to-end.

\section{Conclusion}
In this paper, we extend a popular generative document model \cite{miao2016neural} built on the variational autoencoder framework.
We explicitly model the local context to separate syntax from the semantic document representations.
This is different from most state-of-the-art document models that only make use of global context, i.e. bag-of-words. In contrast to \cite{le2014distributed}, we provide a probabilistic model that does not need to be trained at prediction time. We compared against several document model baselines on established tasks and found that our model consistently finds better representations, produces more topical clusters and is more robust to syntactic peculiarities of the training data.

There are several promising directions for future inquiry. The variational autoencoder framework leaves room for much creativity. Since our extension is independent of the exact encoder and decoder, future work may readily combine it with more flexible encoders or decoders with higher capacity. Moreover, the idea of explaining away local context by explicitly modeling it is sufficiently general to be applicable in other models and for purposes other than document modeling.

% \section*{Acknowledgments}
% \input{acknowledgments}

%\bibliography{acl2018}
\bibliography{acl2018}
\bibliographystyle{acl_natbib}

%\newpage
%\appendix
%\input{appendix}

\end{document}